\documentclass[11pt,a4paper]{article}
\usepackage[hyperref]{acl2021}
\usepackage{times}
\usepackage{latexsym}

\usepackage{microtype}

\aclfinalcopy 
\def\aclpaperid{2968} 

\title{Instructions for ACL-IJCNLP 2021 Proceedings}

\author{First Author \\
  Affiliation / Address line 1 \\
  Affiliation / Address line 2 \\
  Affiliation / Address line 3 \\
  \texttt{email@domain} \\\And
  Second Author \\
  Affiliation / Address line 1 \\
  Affiliation / Address line 2 \\
  Affiliation / Address line 3 \\
  \texttt{email@domain} \\}

\date{}

\usepackage{epsfig}
\usepackage{graphicx}
\usepackage{amsmath}
\usepackage{amssymb}
\usepackage{enumitem}
\usepackage{comment}
\usepackage{algorithm2e}
\usepackage{txfonts}

\newcommand\rowan[1]{{\color{purple}\{\textit{#1}\}$_{rz}$}}

\newcommand\anth[1]{{\color{orange}\{\textit{#1}\}$_{az}$}}

\newcommand\antoine[1]{{\color{red}\{\textit{#1}\}$_{ab}$}}

\newcommand{\datasetname}{EMU}
\newcommand{\taskname}{Edited Media Understanding Frames}
\newcommand{\modelname}{PELICAN}

\newcommand{\imagegptname}{Cross-Modality GPT-2}
\newcommand{\dimension}{type}
\newcommand{\dimensions}{types}

\DeclareMathOperator*{\argmax}{argmax}

\usepackage{microtype}
\usepackage{subcaption}
\usepackage{tabularx}
\usepackage{multirow}
\usepackage{graphicx}
\usepackage{float}
\usepackage{tikz}
\usepackage{tikz-qtree}
\usetikzlibrary{decorations.text, decorations.pathreplacing}
\usepackage{booktabs,array,xcolor,makecell}
\usepackage{microtype}
\usepackage{hyphenat}

\usepackage{physics}

\usepackage{soul}
\usepackage{booktabs}
\usepackage{relsize}
\usepackage{colortbl,makecell}

\usepackage[normalem]{ulem}
\useunder{\uline}{\ul}{}

\renewcommand*{\mathellipsis}{%
  \mathinner{{\ldotp}{\ldotp}{\ldotp}}%
}

\definecolor{bkgray}{gray}{0.8}

\newcommand{\subjectone}{{\small{\texttt{subject1}}}}

\newcommand{\subjecttwo}{{\small{\texttt{subject2}}}}

\newcommand{\subjectthree}{{\small{\texttt{subject3}}}}
\newcommand{\subjectX}{{\smaller{\texttt{subjectX}}}}
\newcommand{\psbattles}{{\smaller\tt \href{https://reddit.com/r/photoshopbattles}{r/photoshopbattles}}}

\makeatletter
\newcommand*{\@rowstyle}{}
\newcommand*{\rowstyle}[1]{
  \gdef\@rowstyle{#1}%
  \@rowstyle\ignorespaces%
}
\newcolumntype{=}{
  >{\gdef\@rowstyle{}}%
}
\newcolumntype{+}{
  >{\@rowstyle}%
}

\begin{document}
\definecolor{amber}{rgb}{1.0, 0.75, 0.0}
\definecolor{midnightblue}{rgb}{0.1, 0.1, 0.44}

\title{Edited Media Understanding Frames: Reasoning About the \\ 
Intents and Implications 
of Visual Disinformation}

\author{Jeff Da$^\spadesuit$ \: \: 
  Maxwell Forbes$^{\spadesuit\heartsuit}$ \: \:
  Rowan Zellers$^{\spadesuit\heartsuit}$ \: \:
  Anthony Zheng$^{\clubsuit}$ \: \: \\
  \textbf{Jena D. Hwang$^{\spadesuit}$ \: \:
  Antoine Bosselut$^{\spadesuit\diamondsuit}$ \: \:
  Yejin Choi$^{\spadesuit\heartsuit}$}\\
  $^\spadesuit$Allen Institute for Artificial Intelligence
  $^\clubsuit$University of Michigan \\
  $^\heartsuit$Paul G. Allen School of Computer Science \& Engineering, University of Washington \\
  $^\diamondsuit$Stanford University \\
  \\ \texttt{\href{http://jeffda.com/edited-media-understanding}{\textbf{\color{midnightblue}jeffda.com/edited-media-understanding}}}
  }

\maketitle

\begin{abstract}


%

Understanding manipulated media, from automatically generated `deepfakes' to manually edited ones, raises novel research challenges. Because the vast majority of edited or manipulated images are benign, such as photoshopped images for visual enhancements, the key challenge is to understand the complex layers of underlying intents of media edits and their implications with respect to disinformation. 

In this paper, we study \textbf{Edited Media Understanding Frames}, a new conceptual formalism to understand visual media manipulation as structured annotations with respect to the intents, emotional reactions, effects on individuals, and the overall implications of disinformation. We introduce a dataset for our task, \datasetname, with 56k question-answer pairs written in rich natural language. We evaluate a wide variety of vision-and-language models for our task, and introduce a new model \modelname, which builds upon recent progress in pretrained multimodal representations. Our model obtains promising results on our dataset, with humans rating its answers as accurate 48.2\% of the time. At the same time, there is still much work to be done -- and we provide analysis that highlights areas for further progress.

\end{abstract}
\newcommand\blfootnote[1]{%
  \begingroup
  \renewcommand\thefootnote{}\footnote{#1}%
  \addtocounter{footnote}{-1}%
  \endgroup
}

\section{Introduction}

\begin{figure}[t]
\centering
  \includegraphics[width=\columnwidth]{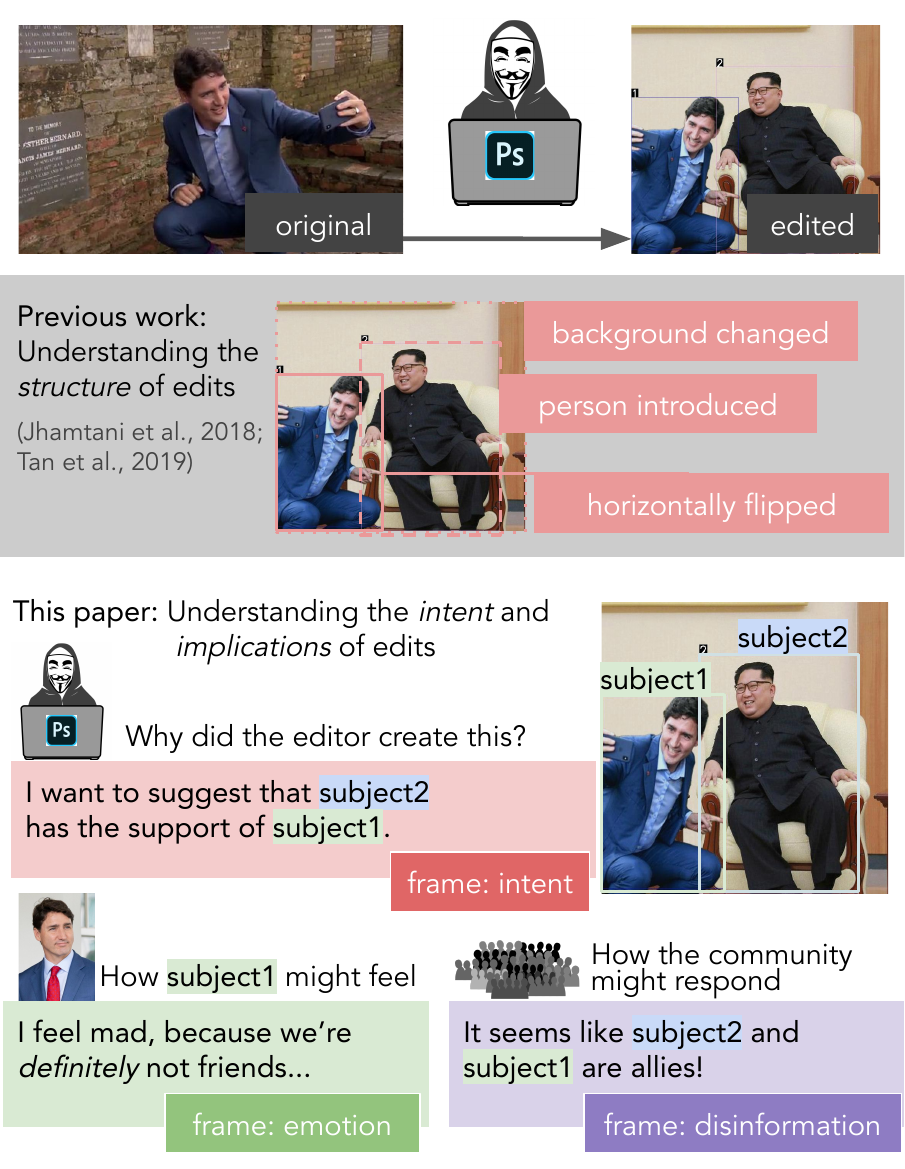}
  \caption{\textbf{Edited Media Understanding Frames.} Given a manipulated image and its source, a model must generate natural language answers to a set of open-ended questions. Our questions test the understanding of the \emph{what} and \emph{why} behind important changes in the image -- like that \subjectone~appears to be on good terms with \subjecttwo.}
  \label{fig:fig1}
\end{figure}

The modern ubiquity of powerful image-editing software has led to a variety of new 
disinformation threats. From AI-enabled ``deepfakes'' to low-skilled ``cheapfakes,'' attackers edit media to engage in a variety of harmful behaviors, such as spreading disinformation, creating revenge porn, and committing fraud \citep[c.f.]{Paris2019DeepfakesAC, chesney2019deep, kietzmann2020deepfakes}. Accordingly, we argue that it is important to develop systems to help spot harmful manipulated media. The rapid growth and virality of social media requires as such, especially as social media trends towards visual content \cite{Gretzel2017TheVT}.

Identifying \emph{whether} an image or video has been digitally altered (i.e., ``digital forgery detection'') has been a long-standing problem in the computer vision and media forensics communities. This has enabled the development of a suite of detection approaches, such as analyzing pixel-level statistics and compression artifacts \cite{farid2009survey, bianchi2012image, bappy2017exploiting} or identifying ``what" the edit was \cite{Tan2019ExpressingVR}.

However, little work has been done on ``why" an edit is made, which is necessary for identifying harm.
Darkening someone’s skin in a family photo because background light made them seem quite pale is generally harmless. 
While such color re-balancing is common, 
darkening Barack Obama’s (or Rafael Warnock’s) skin in campaign ads was clearly meant as a harmful edit by the editor that did it.\footnote{\href{https://www.politico.com/news/2020/12/05/georgia-senate-old-new-south-442423}{How Georgia’s Senate race pits the Old South against the New South. https://www.politico.com/news/2020/12/05 /georgia-senate-old-new-south-442423}} 
We choose to focus on the ``why'' -- we define a schema for approaching the problem of intent and provide a rich set of natural language responses. We also make a significant contribution towards the ``what:" we include a physical-change question, provide rationales based in physical changes, and give structured annotations (bounding boxes) on what was changed in the edit.

We introduce Edited Media Understanding Frames~(\datasetname), a new conceptual formalism that captures the notions of ``why" and ``what" in image editing for language and vision systems (Figure \ref{fig:fig1}). Following literature on pragmatic frames \cite{Sap2017ConnotationFO,Sap2020SocialBF,Forbes2020SocialC1}---derived from frame semantics \cite{Baker1998TheBF}--- we formalize EMU frames along six dimensions that cover a diverse range of inferences necessary to fully capture the scope of visual disinformation. We delve into the concept of \textit{intention} as discussed by the fake news literature \cite{rashkin2017truth, shu2017fake, zhou2020survey} to capture editor's \textit{intent} such as motivation for edit and intent to deceive, as well as the resulting \textit{implications} of the edited content.
For every dimension we collect both a classification label and a free-form text explanation. For example, for frame \textit{intent}, a model must classify the intent of the edit, and describe why this classification is selected.

We then introduce a new dataset for our task, \datasetname, with 56k annotations over 8k image pairs. To kickstart progress on our task, we introduce a new language and vision model, \modelname, that leverages recent progress in pretrained multimodal representations of images and text \cite{Tan2019LXMERTLC, lu2019vilbert, li2019visualbert}. We compare our model to a suite of strong baselines, including a standard VLP model \cite{Zhou2019UnifiedVP}, and show key improvement in terms of ability to reason about co-referent subjects in the edit. Nevertheless, our task is far from solved: a significant gap remains between the best machine and human accuracy.

Our contributions are thus as follows. First, we introduce a new task of \taskname, which requires a deep understanding of \emph{why} an image was edited, and a corresponding dataset, \datasetname, with 56k captions that cover diverse inferences. In addition, we introduce a new model, \modelname, improving over competitive language-and-vision transformer baselines. Our empirical study demonstrates promising results, but significant headroom remains. We release our dataset at \texttt{jeffda.com/edited-media-understanding} to encourage further study in discovering pragmatic markers of disinformation.

 \section{Defining Edited Media Understanding Frames}
\label{sec:ourtask}

\begin{figure*}[!htb]
\centering
  \includegraphics[width=\textwidth]{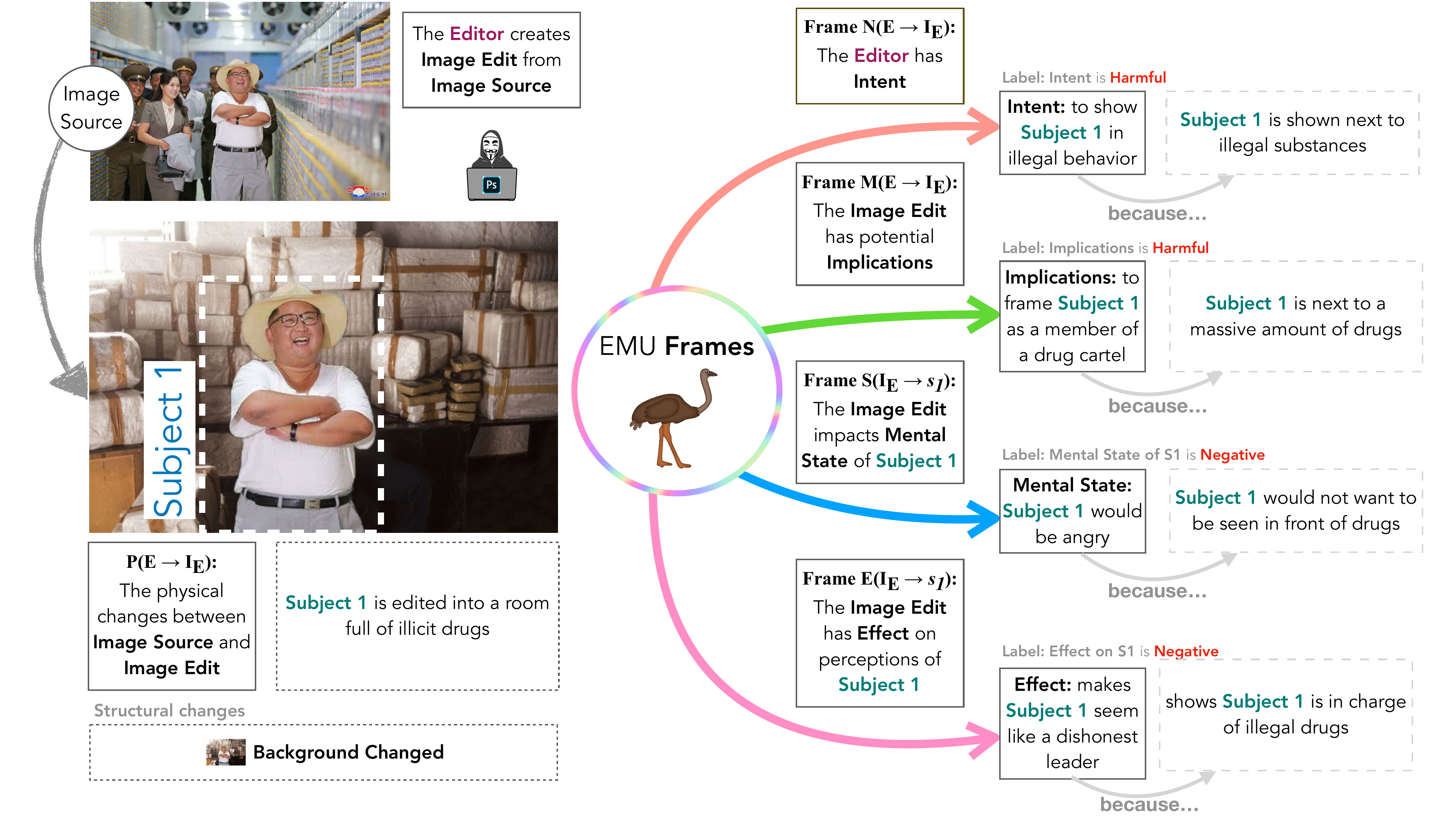}
  \caption{An example from \datasetname. Given a source image and its edit, and a list of main subjects in the image, we collect a label $\mathbf{l}$ and natural language responses (reponse to frame $\mathbf{y}$ and rationale $\mathbf{r}$ to applicable open-ended questions $\mathbf{q}$ covering each of five frames $f \in \mathcal{F}$. We also collect structural annotations $\mathbf{a_i}$ highlighting the edited sections of the image.}
  \label{fig:summary}
\end{figure*}

Through an edit \textit{e} on source image \textit{i} (e.g. ``\textit{e} $=$ \textit{x} is edited into a room full of drugs"), an editor can cause harm to the subject \textit{x}'s \textit{mental state} (\textit{mental state}: ``\textit{x} is angry about \textit{e}") and \textit{effect} \textit{x}'s image (\textit{effect}: ``\textit{e} makes \textit{x} seem dishonest") \cite{Rashkin2016ConnotationFA}. The editor does this through the \textit{intention} of the edit (\textit{intent}: ``\textit{e} intends to harm \textit{x}'s image") and changing the \textit{implications} of the image (\textit{implication}: ``\textit{e} frames \textit{x} as a drug cartel member") \cite{Forbes2020SocialC1, Sap2020SocialBF, Paris2019DeepfakesAC}.

To this end, we collect edits $e$ and source images $i$ from Reddit's \psbattles~community. There is no readily available (large) central database of harmful image edits, but \psbattles~is replete with suitable complex and culturally implicative edits (e.g., reference to politics or pop culture). This provides us with relevant image edits at a reasonable cost without advocation for dangerous training on \textit{real} harmful image edits. Keeping the source image $i$ in the task allows us to sustain the tractability of the image edit problem \cite{Tan2019ExpressingVR, Jhamtani2018LearningTD}.
\subsection{Edited Media Understanding Frames: Task Summary}
Given an edit $e$: $I_S \rightarrow I_E$, we define an edited media understanding frame $\mathcal{F}(*)$ as a collection of typed dimensions and their polarity assignments: (i) \textbf{physical} $P(I_S \rightarrow I_E)$: the changes from $I_S \rightarrow I_E$, (ii) \textbf{intent} $N(E \rightarrow I_E)$: whether the Editor $E$ implied malicious intent in $I_S \rightarrow I_E$, (iii) \textbf{implication} $M(E \rightarrow I_E)$: how $E$ might use $I_E$ to mislead, (iv) \textbf{mental state} $S(I_E \rightarrow s_i)$: whether the predicate $I_E$ impacts the emotion of a role $s_i$, (v) \textbf{effect} $E(I_E \rightarrow s_i)$: the effect of $I_E$ on $s_i$. We assume frames can be categorized as harmful or not harmful with polarity $\mathbf{l}$ $\in \{+, -\}$. Each polarity $\textbf{l}$ can be interpreted with \textit{reason} $\mathbf{y}$, and that each reason can be supported with \textit{rationale} $\mathbf{r}$.

Technically, a model is given the following as input:

\begin{itemize}[wide, labelwidth=!,listparindent=0pt, labelindent=0pt,noitemsep,topsep=1pt,parsep=2pt,leftmargin =*]
\item A source image $I_S$, and an edited image $I_E$.
\item A list of important subjects: expressed as bounding boxes $\mathbf{b}_i$ for each subject.
\item An open-ended question $\boldsymbol{q}$ associated with $~\mathcal{F}(*)$; e.g., ``How might \subjectthree~feel upon seeing this edit?''
\item A list of annotated boxes $\mathbf{a}_i \in I_E$ marking the objects in the image that were \textit{introduced} and \textit{modified}, and a true/false label denoting if the background was changed.
\end{itemize}

A model must produce the polarity classification $\mathbf{l'} \in \{+, -\}$, interpretation of the polarity (response $\mathbf{y'}$) and rationale for interpolation $\mathbf{r'}$. (For the physical frame, only $\textit{y}$ needs to be generated). Figure \ref{fig:summary} shows an example of our task configuration. The lexicon of the label is fixed for each $F(*)$ (e.g. for $N(*)$, $- \rightarrow$ harmful, $+ \rightarrow$ harmless).

\section{\datasetname: A Corpus of Edited Media Understanding Frames}

\paragraph{Sourcing Image Edits} We source our image edits from the \psbattles~community on Reddit which hosts regular Photoshop competitions, where given a \textit{source photo}, members submit a comment with their own \textit{edited photo}.

We collect 8K image edit pairs (source and edited photo pairs) from this community by, first, manually curating a list of more than 100 terms describing people frequently appearing in Photoshop battles posts. Then, we screen over 100k posts for titles that contain one or more of these search terms resulting in 20k collected image pairs. Additionally, we run an object detector \cite{He2017MaskR} to ensure that is at least one person present in each image as a means for ensuring that annotators do not see image pairs without any subjects.

\begin{table}[]
\centering\frenchspacing\small{\renewcommand{\arraystretch}{1.1}
\begin{tabular}{@{} p{1.7cm}@{\hspace{0.4em}} p{1.5cm} p{3.9cm} @{\hspace{0.4em}} l @{}} \toprule
{\smaller \textsc{Frame}} & Notation & Related Question \\ \hline
{\smaller \textsc{Physical}} & $P(I_S \rightarrow I_E)$ & What changed in this image edit? \\
{\smaller \textsc{Intent}} & $N(E \rightarrow I_E)$ & Why would someone create this edit? \\
{\smaller \textsc{Implication}} & $M(E \rightarrow I_E)$ & How might this edit be used to mislead? \\
{\smaller \textsc{Mental State} [of \subjectX]} & $S(I_E \rightarrow s_i)$ & How might this image edit make \subjectX~feel? \\
{\smaller \textsc{Effect} [on \subjectX]} & $E(I_E \rightarrow s_i)$ &  How could this edit mislead public perception of \subjectX? \\
 \bottomrule
\end{tabular}}
\caption{Questions for each of the frames in Edited Media Understanding Frames. Each frame is associated with a question that allows human annotators to address the frame, and models to generate $l, y, r$ for the given frame.}
\label{fig:dimdis}
\end{table} 
\paragraph{Annotating Image Edits}
We ask a group of vetted crowd workers to identify the main subjects in an image edit and answer open-ended questions in natural language. Each image is annotated by 3 independent crowd workers.

\begin{figure}[t]
\centering
  \includegraphics[width=0.5\textwidth]{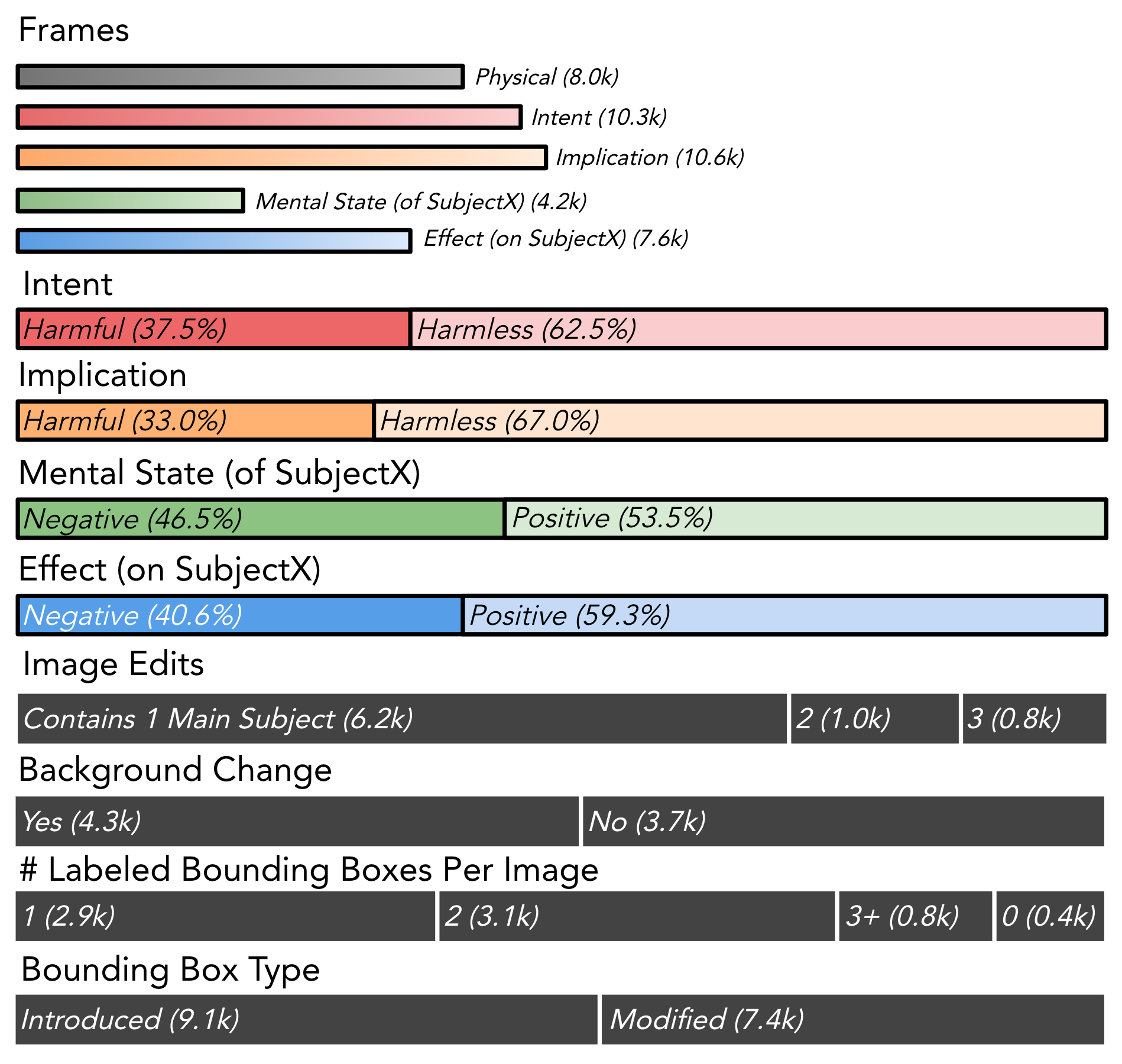}
  \caption{Statistics for EMU. We consider five question types, which in aggregate require a strong understanding of the image edit. The first three types are subject agnostic, though annotations refer explicitly to subjects through subject tags; two (with \subjectX) are subject-specific.}
  \label{fig:stats}
\end{figure}

Crowd workers are first presented with a numbered set of people bounding boxes (produced by Mask R-CNN \cite{He2017MaskR}) over the edited image and are asked to \textbf{select subjects} that are significant to the edits (as opposed, say, a crowd in the background). Once subjects are selected, the annotators are asked to assign \textbf{classification labels} for each of the five possible question types and provide \textbf{free-form text} answers for each question (when applicable). For the classification label, we retain the majority vote (Fleiss $\kappa$ $= 0.67$).
%
%
In a separate and final pass, we explicitly identify which portions of the modified image is \textbf{\textit{introduced}} or \textbf{\textit{altered}} by asking the workers to to label the most important sections of the modified image and selecting one of the two labels. The statistics of the dataset are shown in Figure \ref{fig:stats}. 

\section{Modeling \taskname}
In this section, we present a new model for \taskname, with a goal of kickstarting research on this challenging problem. As described in Section~\ref{sec:ourtask}, our task differs from many standard vision-and-language tasks both in terms of format and required reasoning: a model must take as input two images (a source image and its edit), with a significant change of implication added by the editor. A model must be able to answer questions, grounded in the main subjects of the image, describing these changes. The answers are either boolean labels, or open-ended natural language -- including explainable rationales.

\subsection{Our model:  \modelname}

\begin{figure*}[t]
\centering
  \includegraphics[width=0.98\textwidth]{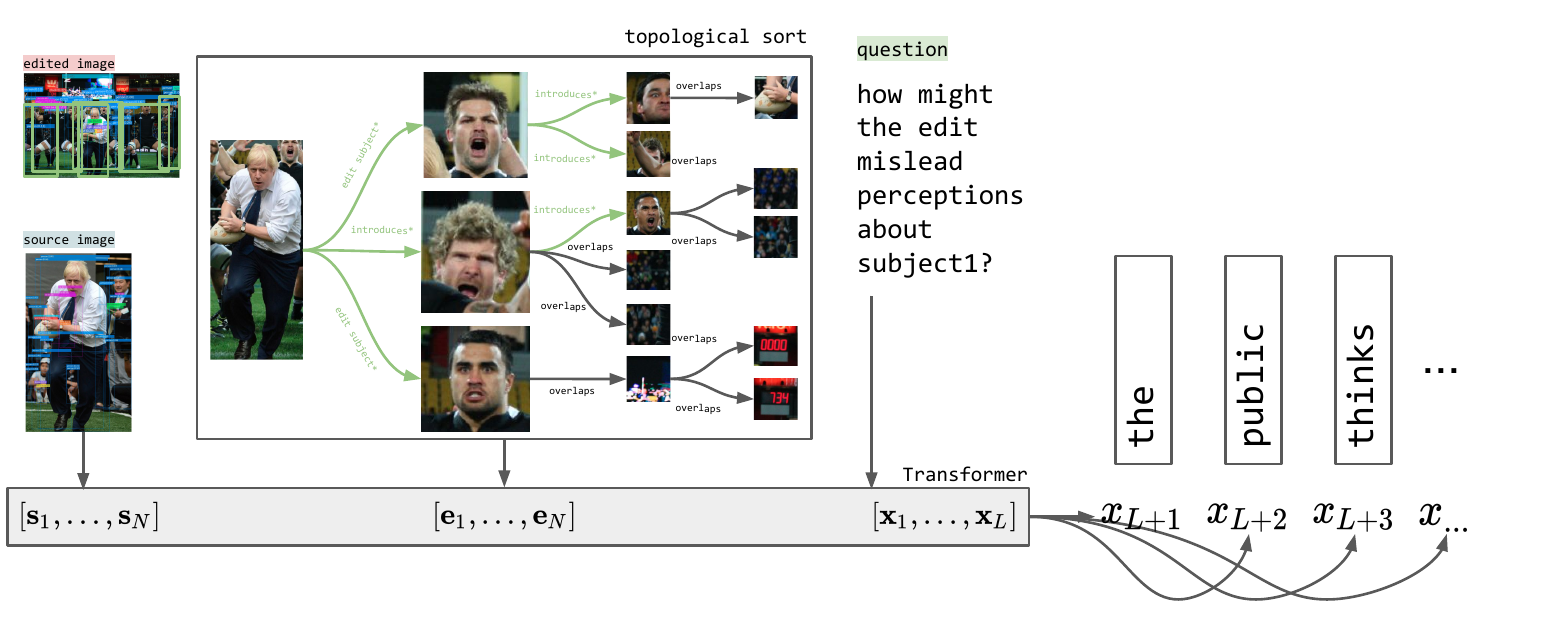}
  \caption{Overview of \modelname. Our model takes as input all regions $\textbf{s}$ from the source image and $\textbf{e}$ from the edited image. We order the regions in $\textbf{e}$ using a topological sort of overlapping boxes, rooted at \subjectone. The green regions marked with an asterisk are additional regions that were introduced, and were labeled through annotators. This ordering allows the model to selectively attend to important image regions in generating an answer to the visual question about \subjectone.}
  \label{fig:generationexamples}
\end{figure*}

For \taskname, \emph{not all image regions are created equal}. Not only is the subject referred to in the question (e.g. \subjectone) likely important, so too are all of the regions in the image edit that are \emph{introduced} or \emph{altered}. We propose to use the annotations that collected for these regions as additional signal for the model to highlight where to attend.\footnote{These annotations are collected from workers, but in theory, it would be possible to train a model to annotate regions as such. To make our task as accessible and easy-to-study as possible, however, we use the provided labels in place of a separate model however.} Not only should a model likely attend to these important regions, it should prioritize attending to regions \emph{nearby} (such as objects that an edited person is interacting with).

We propose to model the (likely) importance of an image region through graph propagation. We will build a directed graph with all regions of the image, rooted at a subject mentioned by the question (e.g. \subjectone). We will then \emph{topologically sort} this graph; each region is then given an embedding corresponding to its sorted position -- similar to the position embedding in a Transformer. This will allow the model to selectively attend to important image regions in the image edit. We use a different position embedding for the image source, and do not perform the graph propagation here (as we do not have \emph{introduced} or \emph{altered} annotations); this separate embedding captures the inductive bias that the edited is more important than the source.

\subsection{Model details and Transformer integration}
In this section, we describe integrating our \emph{importance embeddings} with a multimodal transformer.

Let the source image be $I_S$ and $I_E$. We use the backbone feature extractor $\phi$ ( Faster-RCNN feature extractor \cite{ren2015faster, anderson2018bottom} to extract $N$ regions of interest for each region:
\begin{equation}
\left[ \mathbf{s}_1, \dots, \mathbf{s}_N \right] = \phi(I_S) \qquad \left[ \mathbf{e}_1, \dots, \mathbf{e}_N \right] = \phi(I_E).
\end{equation}
We note that some of these regions in $\mathbf{e}_1, \dots, \mathbf{e}_N$ are provided to the model (as annotated regions in the image); the rest are detected by $\phi$. These, plus the language representation of the question, are passed to the Transformer backbone $T$:
\begin{equation}
[\mathbf{z}_{1}\mathellipsis\mathbf{z}_{N+L}] = T([\mathbf{s}_1 \mathellipsis \mathbf{s}_N], \left[ \mathbf{e}_1, \dots, \mathbf{e}_N \right], [\mathbf{x}_1 \mathellipsis \mathbf{x}_L])
\end{equation}

Important for \datasetname, $\mathbf{z}_{2N+1}, \ldots, \mathbf{z}_{2N+L}$ serve as language representations. Training under a left-to-right language modeling objective, we can predict the next \emph{next token} $x_{L+1}$ using the representation $\mathbf{z}_{N+L}$.

\subsubsection{Prioritization Embeddings from Topological Sort}

Transformers require \emph{position embeddings} to be added to each image region and word -- enabling it to distinguish which region is which. We supplement the position embeddings of the regions $\{\mathbf{e}_1 ... \mathbf{e}_N\}$ in the edited image $I_E$ with the result of a topological sort.

\textbf{Graph definition.} We define the graph over image regions in the edited image as follows.  We begin by sourcing a seed region $\mathbf{s} \in \{\mathbf{e}_1 ... \mathbf{e}_N\}$. Let $G = (V, E)$, where each $v \in V$ represents metadata of some $\mathbf{r}_i \in \phi(I_E)$, defined as $v_i \in m(I_E)$ for simplicity, s.t.:

\begin{equation}
    v_i = \{x_1, y_1, x_2, y_2, s_i, l_i\}
\end{equation}

where $x_1, y_1, x_2, y_1$ represents the bounding box of $\mathbf{r_i}$, $s_i \in \{1, 0\}$ denoting if $r_i$ is a subject of $I_E$, and $l_i \in \{introduced,~altered\}$ denoting the label of $r_i$.

We build the graph iteratively: for each iteration, we define an edge $\mathbf{e} = \{v, u\}; u \in V$ s.t.:
\begin{equation}
    \forall v \in m(I_E), \forall u \in V, E = E \cup (u, v) \in E'
\end{equation}
We define $E'$ as the set of edges $(u, v)$ in which $u$ and $v$ are \textit{notationally similar}. We define three cases in which this is true: if $s_i \in u_i \land s_j \in v_j$, if $l_i \in u_i = l_j \in v_j$, and if $x_1, y_1, x_2, y_2 \in u_i$ and $x_3, y_3, x_4, y_4 \in u_i$ overlaps, in which the percentage overlap is defined by standard intersection-over-union:
\begin{equation}
\frac{\min\{x_4, x_2\} - \max\{x_3, x_1\}}{\min\{y_4, y_2\} - \max\{y_3, y_1\}}
\end{equation}

We cap the number of outgoing edges at 3, and prevent cycles by allowing edges only to unseen image regions. In cases where there are more than three possible edges, we add edges in the order defined in the previous paragraph, and break overlap ties via maximum overlap.

To produce embeddings, we run topological sort over the directed graph to assign each image region an embedding, then assign an embedding to each image region based on the ordered index. The embedding is zeroed out for image regions that are missing from the DAG, and from the source image (which are unlabeled). We include bounding box and class labels. To generate text and classification labels, we attach the embeddings onto the input for an encoder-decoder structure.

\section{Experimental Results on \datasetname}
\label{sec:results}

\begin{figure*}[t]
\centering
  \includegraphics[width=0.98\textwidth]{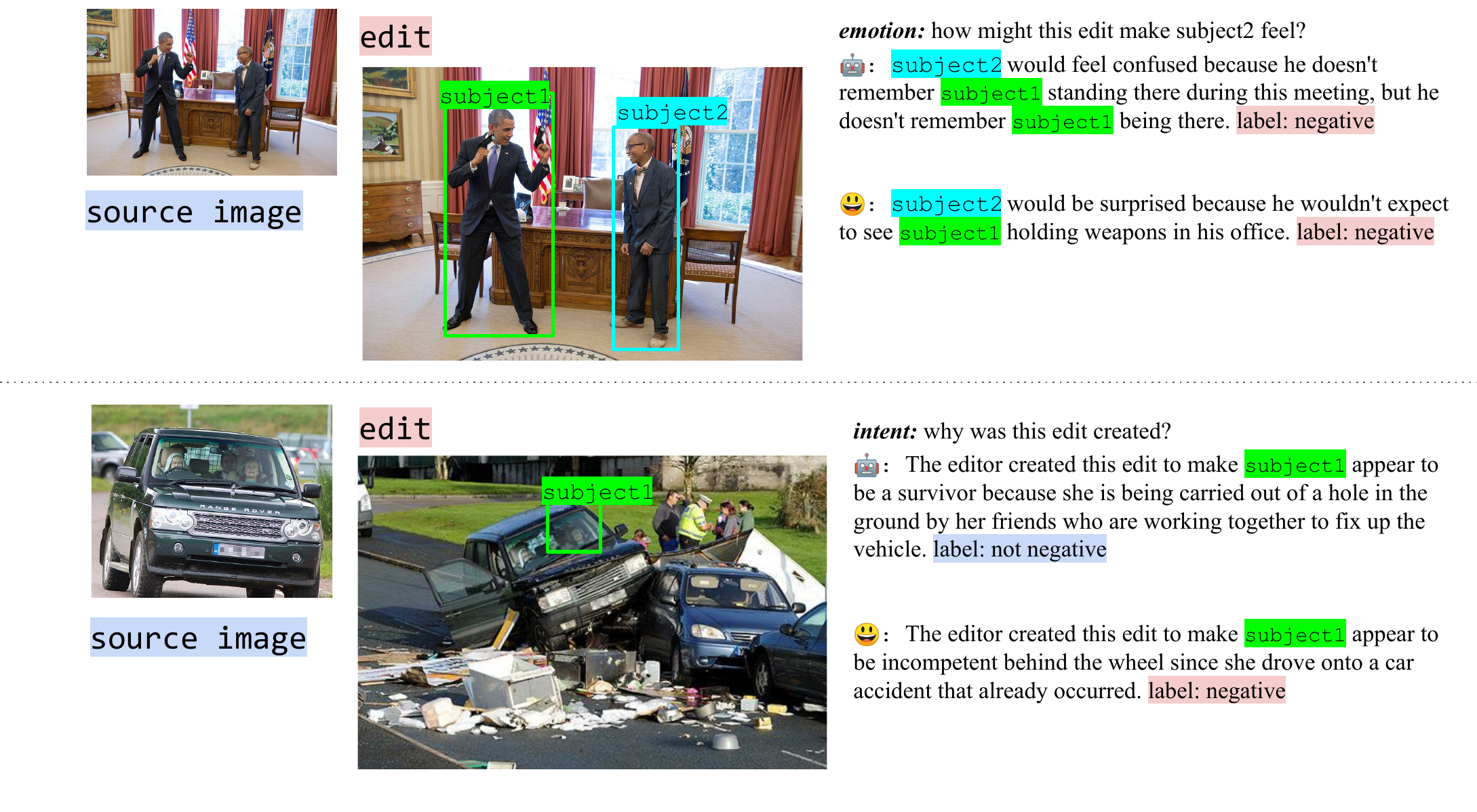}
  \caption{Generation examples from \modelname, marked with results from human evaluation. \modelname~is able to correctly reference marked figures and is able to infer intent accordingly across each question type.}
  \label{fig:generationexamples}
\end{figure*}

In this section, we evaluate a variety of strong vision-and-language generators on \datasetname. Similar to past work on VQA, we rebalance our test set split ensuring a 50/50 split per question type of maliciously labeled captions. We provide two human evaluation metrics -- head-to-head, in which generated responses are compared to human responses, and accuracy, in which humans are asked to label if generated responses are accurate in regards to the given edit.

\subsection{Baselines}

In addition to evaluating \modelname, we compare and evaluate the performance of various potentially high-performing baselines on our task.

\definecolor{lightgray}{rgb}{0.99, 0.99, 0.99}
\newcolumntype{g}{>{\columncolor{lightgray}}c}
\begin{table*}[]
\footnotesize
\centering
\resizebox{\textwidth}{!}{
\begin{tabular}{l||c|c|c|c||c|c}
\multicolumn{1}{c}{} & \multicolumn{4}{g}{{\smaller Automated metrics}} & \multicolumn{2}{g}{{\smaller Human evaluation}} \\
Model & \thead{Perplexity $\downarrow$} & \thead{ROUGE-L $\uparrow$} & \thead{METEOR $\uparrow$} & \thead{Accuracy $\uparrow$} & \thead{Head-to-\\Head $\uparrow$} & \thead{Accurate \% $\uparrow$} \\ \toprule
Humans & n/a & n/a & n/a & 89.8 & 50.0 & 95.2 \\ \cmidrule{1-7}
Retrieval Baseline & n/a & 11.5 & 7.2 & 51.9 & 4.4 & 20.6 \\
GPT-2 \cite{Radford2019LanguageMA} & 26.6 & 10.3 & 6.2 & 50.0 & 0.0 & 3.0 \\
\imagegptname & 22.1 & 12.0 & 7.9 & 51.0 & 4.1 & 10.4 \\
Dynamic RA \cite{Tan2019ExpressingVR} & 23.1 & 13.2 & 8.9 & 51.8 & 5.3 & 12.4 \\
VLP \cite{Zhou2019UnifiedVP} & 12.3 & 18.5 & 10.5 & 53.2 & 9.3 & 20.3 \\
\modelname \space \texttt{REAL} (ours) & 11.6 & 19.5 & 10.8 & 54.1 & 11.3 & 25.5 \\
\modelname \space (ours) & \textbf{11.0} & \textbf{22.1} & \textbf{11.6} & \textbf{55.4} & \textbf{14.6} & \textbf{48.2}
\end{tabular}}
\caption{Experimental results on \datasetname. We compare our model, \modelname, with several strong baseline approaches. We calculate generative metrics (e.g. METEOR) by appending the rationale to the response. PELICAN \texttt{REAL} describes a version of PELICAN trained on EMU without additional human annotation (6.1).}
\label{fig:results}
\end{table*}

\textbf{a. Retrieval}. For a retrieval baseline, which generally performs well for generation-based tasks, we use features from ResNet-158 \cite{He2016DeepRL}, defined as $\phi$, to generate vectors for each $I_E$ in the test set. We then find the most similar edited image $I_T$ in the training set $\mathbf{T}$ via cosine similarity:

\begin{equation}
\argmax_{I_T \in \mathbf{T}} \frac{\phi(I_E) \cdot \phi(I_T)}{\norm{\phi(I_E)} \times \norm{\phi(I_T)}}
\end{equation}

We use the captions associated with the most similar image in the training set.

\textbf{b. GPT-2 \cite{Radford2019LanguageMA}}. As a text-only baseline, we use the 117M parameter model from GPT-2, fine-tuned on the captions from our dataset. Since the images are not taken into consideration, we generate from the seeds associated with each question \dimension \space and use the same captions for all images in the test set.

\textbf{c. \imagegptname}. We test a unified language-and-vision model on our dataset. Similar to \cite{Alberti2019FusionOD}, we append the visual features $\phi(I_S)$ and $\phi(I_E)$ to the beginning of the token embeddings from GPT-2 (117M). For the questions involving a \texttt{subject}, we append an additional vector $\phi(r)$, where $r$ is the region defined by the bounding box for that \texttt{subject}.

\textbf{d. Dynamic Relational Attention \cite{Tan2019ExpressingVR}}. We test the best model from previous work on image edits on our task, Dynamic Relational Attention. We train the model from scratch on our dataset, using the same procedure as \cite{Tan2019ExpressingVR}. We seed each caption with the relevant question.

\textbf{e. VLP \cite{Zhou2019UnifiedVP}}. We test VLP, a pre-trained vision-and-language transformer model. For image captioning, VLP takes a single image as input and uses an off-the-shelf object detector to extract regions, generation a caption using sequence-to-sequence decoding and treating the regions as a sequence of input tokens. 

To generate a caption for a particular question \dimension, we fix the first few generated tokens to match the prefix for that question \dimension. We fine-tune VLP starting from weights pre-trained on Conceptual Captions (3.3m image-caption pairs) \cite{Sharma2018ConceptualCA} and then further trained on COCO Captions (413k image-caption pairs) \cite{Lin2014MicrosoftCC}.


\subsection{Quantitative Results and Ablation Study}
\definecolor{lightgray}{rgb}{0.99, 0.99, 0.99}
\newcolumntype{g}{>{\columncolor{lightgray}}c}
\begin{table}[]
\footnotesize
\centering
\resizebox{0.5\textwidth}{!}{
\begin{tabular}{l||c||c}
\multicolumn{1}{c}{} &  \multicolumn{1}{g}{{\smaller Auto Eval}} & \multicolumn{1}{g}{{\smaller Human Eval}} \\
Model & \thead{Accuracy $\downarrow$} & \thead{Accuracy $\uparrow$} \\ \toprule
\modelname & \textbf{55.40} & \textbf{48.2} \\ \cmidrule{1-3}
physical & n/a & 60.5 \\
intent & 55.2 & 43.0 \\
implication & 60.1 & 49.9 \\
mental state [of subjectx] & 54.6 & 42.5 \\
effect [on subjectx] & 53.7 & 41.1 \\
\cmidrule{1-3}
$-$ pretraining & 54.6 & 44.0 \\
$-$ annotated features & 54.4 & 40.1 \\
$-$ directed graph & 54.9 & 45.2 \\
$-$ source image & 55.3 & 47.5 \\
\end{tabular}}
\caption{Ablation study for \modelname. We also explore the performance of \modelname \space across each frame type.}
\label{fig:ablations}
\end{table}

We present our results in Table \ref{fig:results}. We calculate generative metrics (e.g. METEOR) by appending the rationale to the response. Generations from \modelname \space are preferred over human generations 14.0\% of the time, with a 0.86 drop in perplexity compared to the next best model. To investigate the performance of the model, we run an ablation study on various modeling attributes, detailed in Table \ref{fig:ablations}. First, we investigate the effect of pretraining (on Conceptual Captions \cite{Sharma2018ConceptualCA, Zhou2019UnifiedVP}). We find that performance drops without pretraining (53.47\%), but surprisingly still beats other baselines. This suggests that the task requires more pragmatic inferences than the semantic learning typically gained from pre-training tasks. Second, we ablate the importance of including annotated ($\mathbf{a_i}$) features from the dataset when creating the directed graph, relying on a seed from a random R-CNN region (54.44\%). We also ablate our use of topological sort and a directed graph by suggesting a simple (but consistent) order for image regions (54.91\%). Finally, we ablate including the visual regions from the source image. The performance is similar (55.35\%), suggesting that \modelname~would be able to perform in real-world settings in which only the edited image is present (e.g. social media posts).

\subsection{Qualitative Results}

Last, we present qualitative examples in Figure \ref{fig:generationexamples}. \modelname \space is able to correctly understand image pairs which require mostly surface level understanding - for example, in the top example, it is able to identify that the gun and action implies negative context, but misunderstands the response with regards to the situation. In the bottom example, we show that \modelname~is able to refer to \subjectone~correctly, but misinterprets the situation to be non-negative.

\section{Future Implications}

\subsection{EMU in the Real World}

To study if EMU is helpful in real-world settings, we train a model of PELICAN on EMU with only the edited image. In this setting, the model must hypothesize which parts of the image were edited and discern the main subjects in the image. At test time, we generate captions for each of the 5 intention-based question types. Results of this version of PELICAN is in Table \ref{fig:results}.

While this evaluation scheme is crude, we find that this version of PELICAN is still able to outperform previous models without usage of the source image. This suggests potential for generations from EMU-trained models in \textit{human-assisted} settings. In an initial human study (given PELICAN \texttt{REAL} captions, classify the edit as disinformation -- were the captions helpful in your decision?) we find that annotators label as helpful 71.5\% of the time. Additionally, annotators tended more often to pick the gold label (89.1\% $\rightarrow$ 95.2\%).

\subsection{Failure Cases in Current Models and Avenues for Future Research}

\begin{figure}[t]
\centering
  \includegraphics[width=0.5\textwidth]{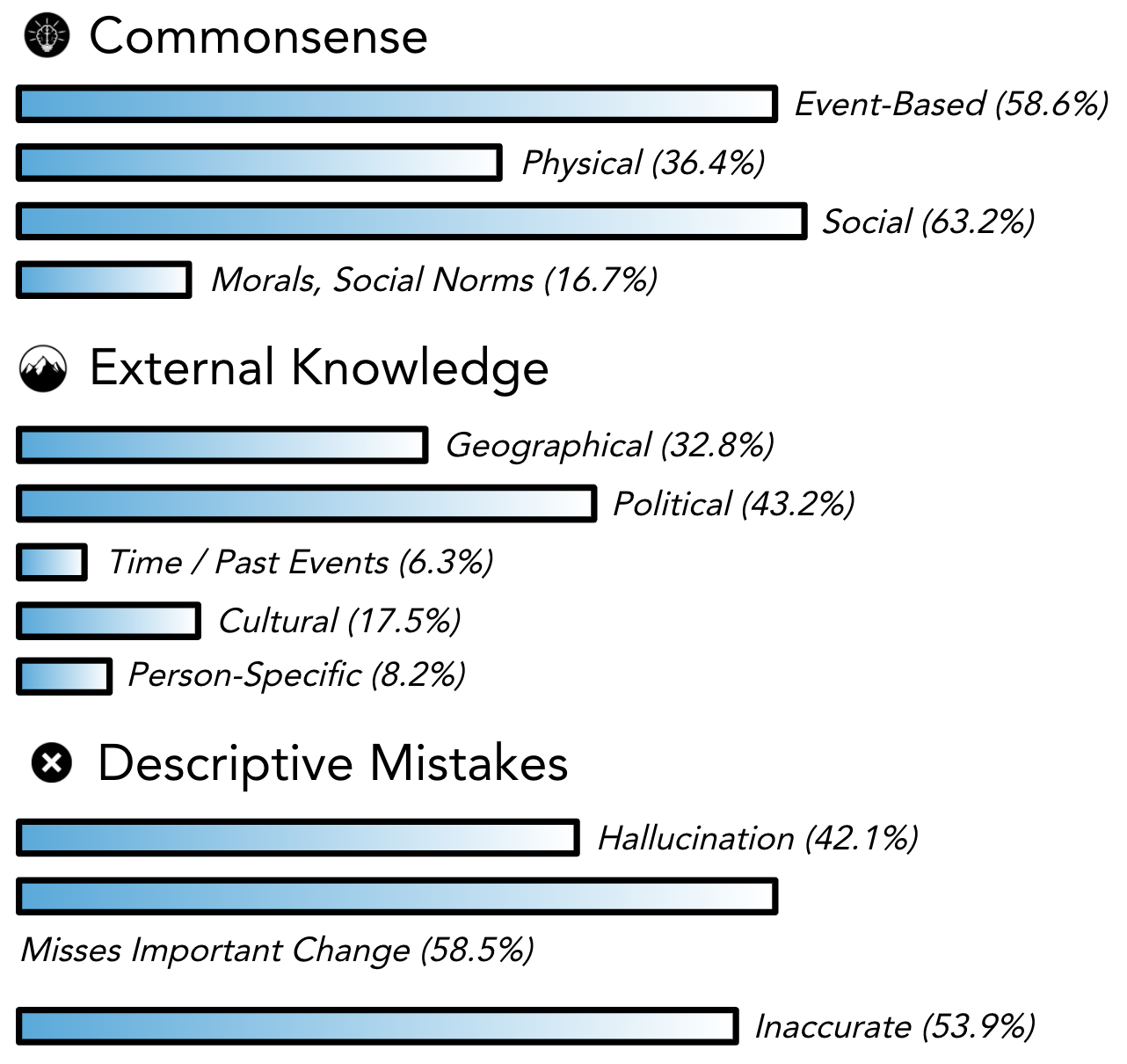}
  \caption{Failure cases from PELICAN, trained on EMU. Commonsense is the largest differentiator between human understanding and model-based analysis of disinformation.}
  \label{fig:mistakes}
\end{figure}

EMU also helps us understand what current vision-and-language models are missing for use on 
disinformation
, by analyzing the reasons and rationales generated. We ask annotators to compare PELICAN-generated captions marked as ``worse" and human captions. Category details are included in the appendix. Figure \ref{fig:mistakes} shows our results. Overall, current models primarily lack the commonsense (event-based and social) to accurately describe disinformation. Geographical (location-based) and political (e.g. knowledge about the job of a president) external knowledge is also a missing component.

PELICAN also still makes mistakes in description-related attributes: describing something other than the important change and an inaccuracy (e.g. wrong color) are the most common. Specific information -- such as information relating to a specific person in the image (i.e. requiring a model to identify the person in the image), and information about a past event -- are the least critical, suggesting that efforts should be focused first on general intelligence rather than named-entity lookup.

\section{Related Work}

\noindent
\textbf{Language-and-Vision Datasets} Datasets involving images and languages cover a variety of tasks, including visual question answering \cite{Agrawal2015VQAVQ, Goyal2017MakingTV}, image caption generation \cite{Lin2014MicrosoftCC, Young2014FromID, Krishna2016VisualGC}, visual storytelling \cite{Park2015ExpressingAI,bosselut2016learning}, machine translation \cite{Elliott2016Multi30KME}, visual reasoning \cite{Johnson2017CLEVRAD, Hudson2019GQAAN, Suhr2019ACF}, and visual common sense \cite{Zellers2019FromRT}.

\textbf{Two-image tasks} Though most computer vision tasks involve single images, some work has been done on exploring image pairs. The NLVR2 dataset \cite{Suhr2019ACF} involves yes-no question answering over image pairs. Neural Naturalist \cite{Forbes2019NeuralNG} tests fine-grained captioning of bird pairs; \cite{Jhamtani2018LearningTD} identifies the difference between two similar images.

\textbf{Image Edits} There has been some computer vision research studying image edits. Unlike our \datasetname~dataset, however, much of this work has focused on modeling lower-level image edits wherein the \emph{cultural implications} do not change significantly between images. For example, \cite{Tan2019ExpressingVR} predicts image editing requests (generate `change the background to blue' from a pair of images). Past work has also studied learning to perform image adjustments (like colorization and enhancement) from a language query \cite{Chen2017LanguageBasedIE, Wang2018LearningTG}. Hateful Meme Challenge \cite{Kiela2020TheHM} is a recent work challenging models to classify a meme as hateful or not.



\section{Conclusion}

We present \taskname -- a language-and-vision task requiring models to answer open-ended questions that capture the intent and implications of an image edit. Our model, \modelname, kickstarts progress on our dataset -- beating all previous models and with humans rating its answers as accurate 48.2\% of the time. At the same time, there is still much work to be done -- and we provide analysis that highlights areas for further progress.
\section*{Acknowledgements}
{
The authors would like to thank Ryan Qiu for help with analysis, and the Amazon Mechanical Turk community for help with annotation. This material is based upon work supported by the National Science Foundation Graduate Research Fellowship under Grant No. DGE1256082, and in part by NSF (IIS-1714566), DARPA CwC through ARO (W911NF15-1-0543), DARPA MCS program through NIWC Pacific (N66001-19-2-4031), NSF (IIS-1714566), and the Allen Institute for AI.
}

\section{Ethical Considerations}

In constructed the EMU dataset, great care was taken to ensure that crowd-workers are compensated fairly for their efforts. To this end, we monitored median HIT completion times for each published batch, adjusting the monetary reward such that at least 80\% of workers always received $>$\$15/hour, which is roughly double the minimum wage in the United States (the country of residence for most Amazon Mechanical Turk workers). This included the qualification and evaluation rounds. The following data sheet summarized relevant aspects of the data collection process \cite{Bender2018DataSF}:

\textsc{A. Curation Rationale}: Selection criteria for the edits included in the presented dataset are discussed Section 3. We selected the highest-rated posts on Reddit, and collected metadata data from annotators marking if the edit is NSFW or offensive.

\textsc{B. Language Variety}: The dataset is available in English, with mainstream US Englishes being the dominant variety, as per the demographic of Amazon Mechanical Turk workers.

\textsc{C. Speaker Demographic}: N/A

\textsc{D. Annotator Demographic}: N/A

\textsc{E. Speech Situation}: All frames were collected and validated over a period of about 12 weeks, between November and January 2020, through the Amazon AMT platform. Workers were given regular, detailed feedback regarding the quality of their submissions and were able to address any questions or comments to the study's main author via Email or Slack.

\textsc{F. Text Characteristics}: In line with the intended purpose of the dataset, the included edits describe social interactions related (but not limited to) platonic and romantic relationships, political situations, as well as cultural and social contexts.

\textsc{G. Recording Quality}: N/A

\textsc{H. Other}: N/A

Lastly, we want to emphasize that our work is strictly scientific in nature, and serves the exploration of machine reasoning alone. It was not developed to offer guidance on misinformation or to train models to classify social posts as misinformation. Consequently, the inclusion of malicious image edits could allow adversaries to train malicious agents to produce visual misinformation. We are aware of this risk, but also want to emphasize that the utility of these agents allow useful negative training signal for minimizing harm that may be cased by agents operating in visual information. It is, therefore, necessary for future work that uses our dataset to specify how the collected examples of both negative and positive misinformation are used, and for what purpose.

\newpage

\newpage

\bibliographystyle{acl_natbib}
\bibliography{acl2021}

\newpage
\newpage

\appendix
\section{Appendices}

\subsection{Reproducibility of Experiments}

We provide downloadable source code of all scripts, and experiments, at \texttt{to-be-provided}. We use two Titan X GPUs to train and evaluate all models, except Dynamic Relational Attention \cite{Tan2019ExpressingVR}, which was trained on a single Titan Xp GPU. For GPT-2 \cite{Radford2019LanguageMA}, we use the 117M parameter model, taking 5 hours to train. Our configuration of VLP \cite{Zhou2019UnifiedVP} has 138,208,324 parameters, taking 6 hours to train. Our model, \modelname, has 138,208,324 parameters, taking 6 hours to train. Our Dynamic Relational Attention model has 55,165,687 parameters, taking 10 hours to train.

\subsection{Reproducibility of Hyperparameters}

For models using GPT-2 as their underlying infrastructure, we use a maximum sequence length of 1024, 12 hidden layers, 12 heads for each attention layer, and 0.1 dropout in all fully connected layers. For Dynamic Relational Attention \cite{Tan2019ExpressingVR}, we use a batch size of 95, hidden dimension size of 512, embedding dimension size of 256, 0.5 dropout, Adam optimizer, and a 1e-4 learning rate. We used early stopping based on the BLEU score on the validation set at the end of every epoch; the test scores reported are for a model trained for 63 epochs. For all models relying on VLP as their underlying infrastructure, we use 30 training epochs, 0.1 warmup proportion, 0.01 weight decay, 64 batch size.

\begin{figure}[t]
\centering
  \includegraphics[width=0.4\textwidth]{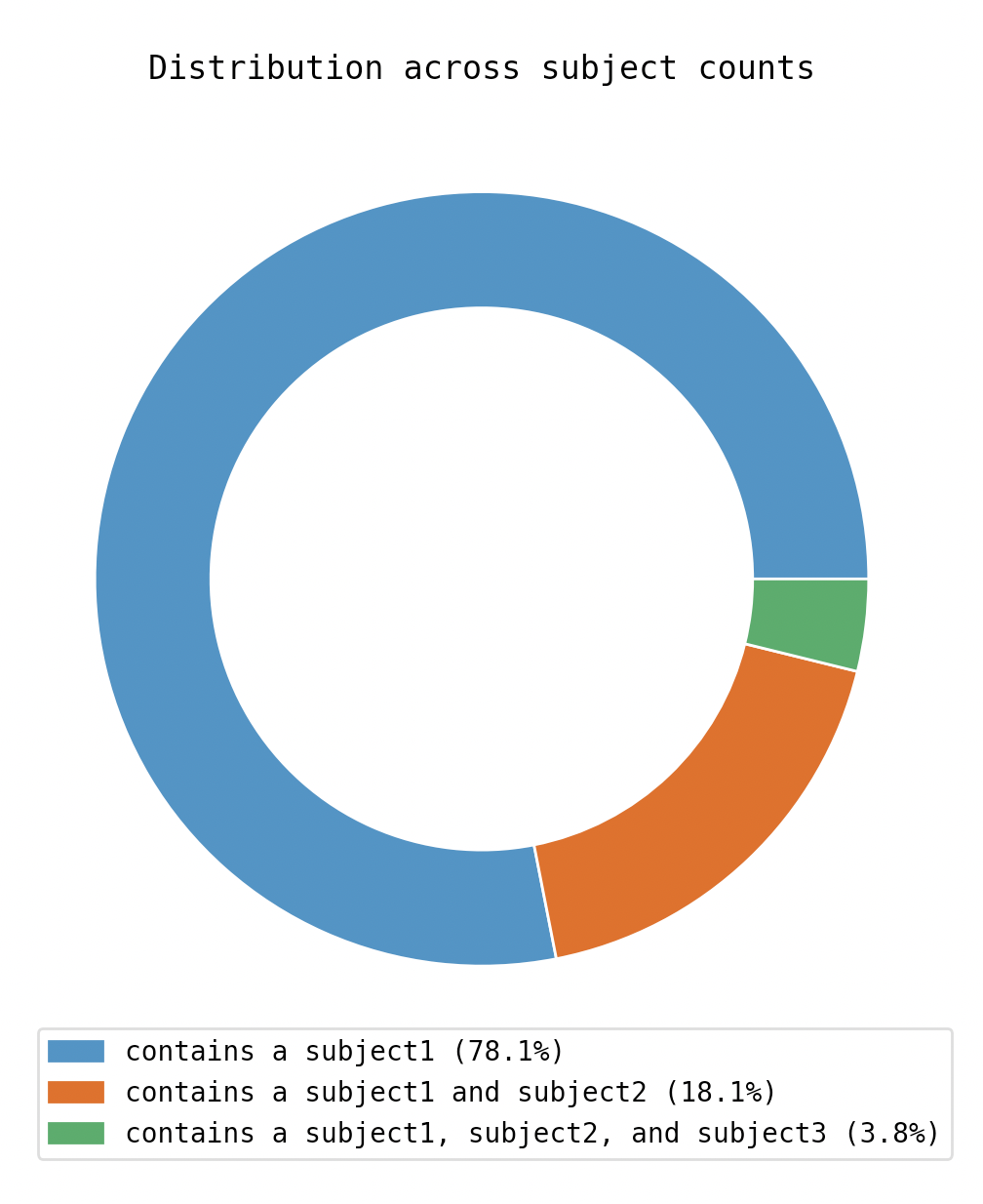}
  \caption{Subject distribution. To highlight our decision for a 3 subject limit, we show that the majority of images contains 1-2 subjects.}
  \label{fig:mpdis}
\end{figure}

\begin{figure}[t]
\centering
  \includegraphics[width=0.4\textwidth]{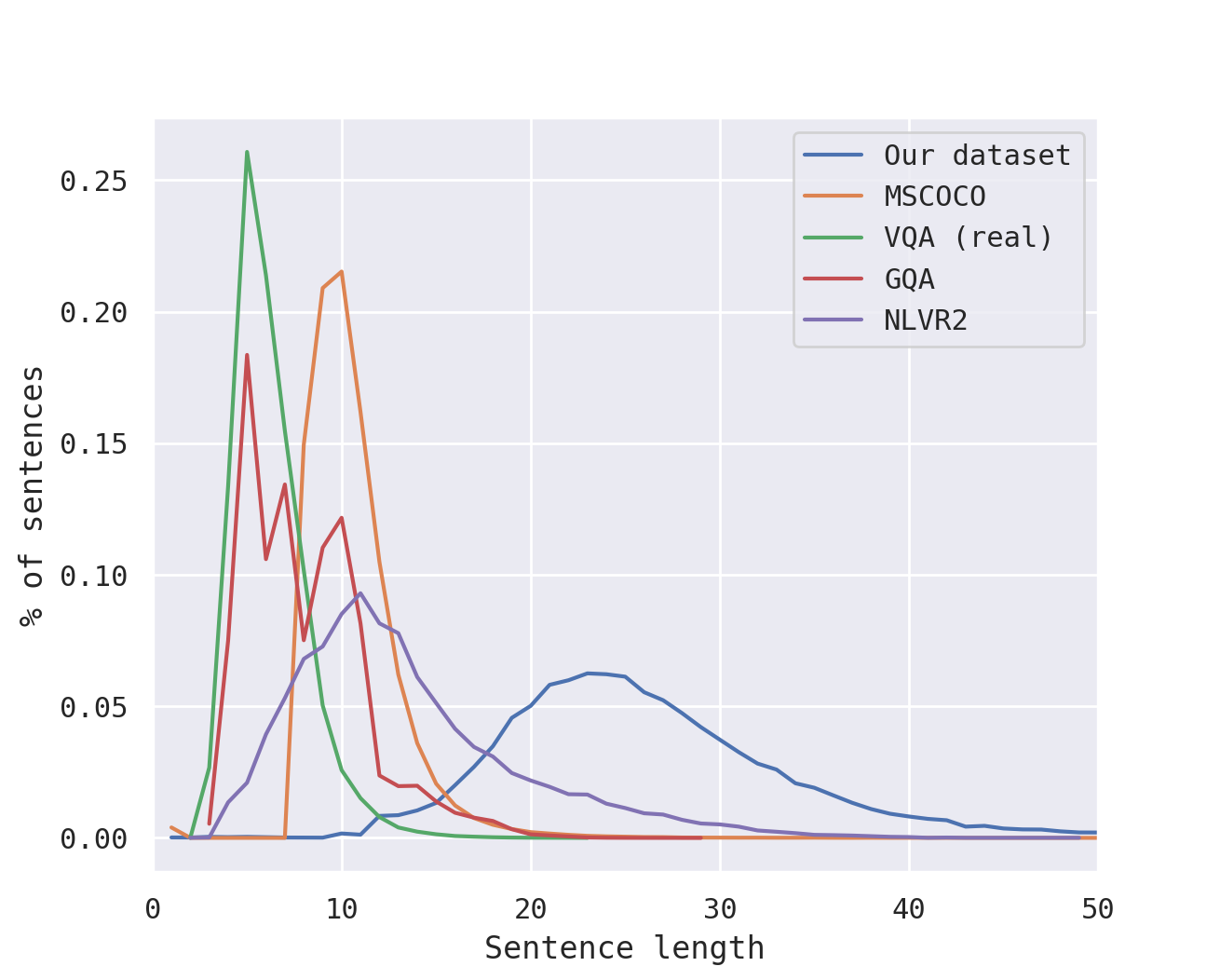}
  \caption{Language sentence length distribution, measured in words, across other language-and-vision datasets. The natural language answers in \datasetname~show a high degree of complexity, with an average sentence length of 26.45 words.}
  \label{fig:length}
\end{figure}

\subsection{Reproducibility of Datasets}

Our dataset has 39338 examples in the training set and 4268 and 3992 examples in the development and test sets respectively. All training on additional datasets (e.g. \cite{Zhou2019UnifiedVP}) matches their implementation exactly. Our train/val/test splits were chosen at random, during the annotation period. No data was excluded, and no additional pre-processing was done. A downloadable link is available at \texttt{to-be-provided} after publication.

\subsection{Data Collection}

\begin{figure}[t]
\centering
  \includegraphics[width=0.4\textwidth]{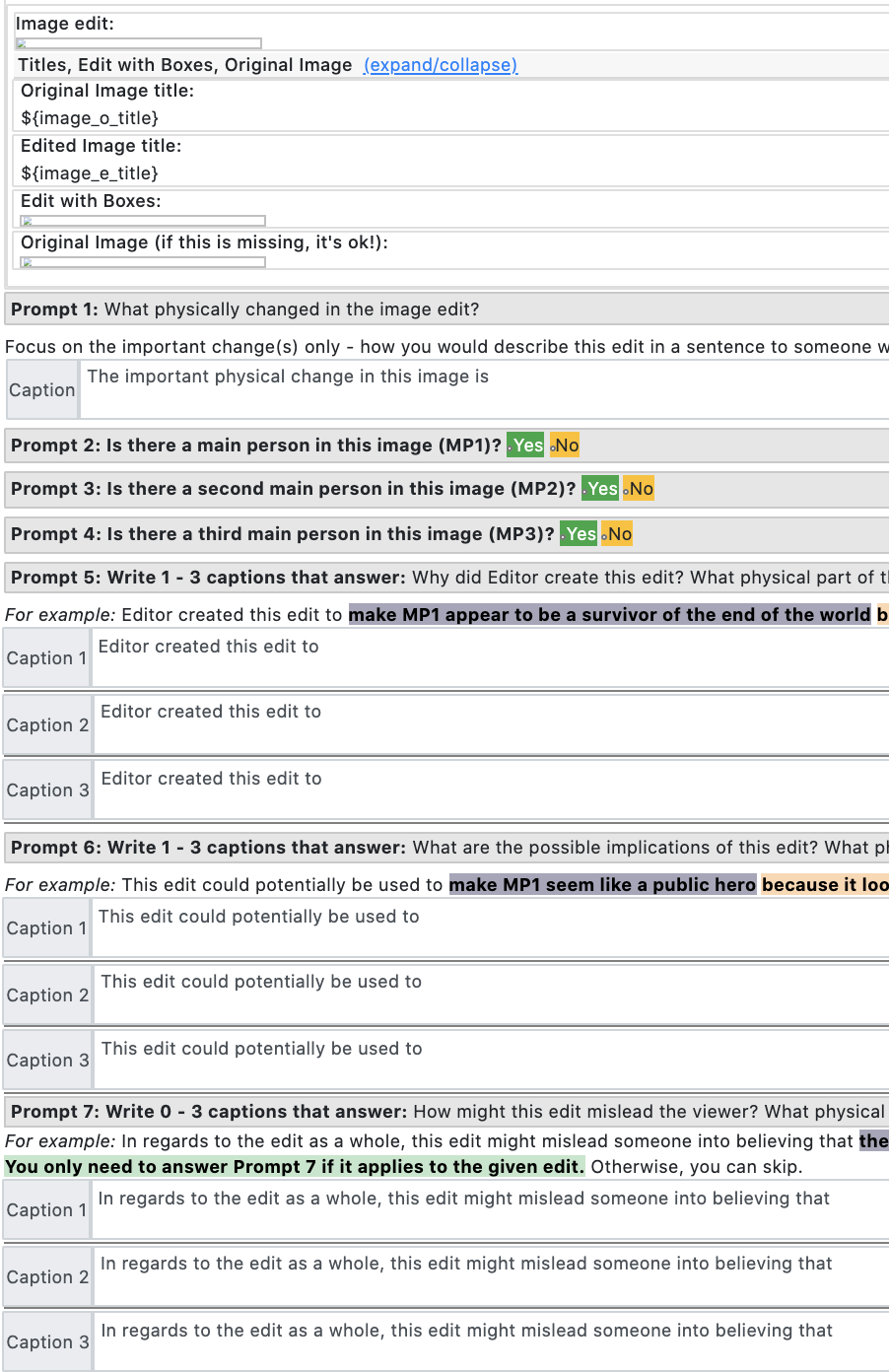}
  \caption{Example of our annotation process.}
  \label{fig:part6}
\end{figure}

For reference and reproducibility, we show the full template used to collect data in Figure \ref{fig:part6}.

\begin{figure}[t]
\centering
  \includegraphics[width=0.4\textwidth]{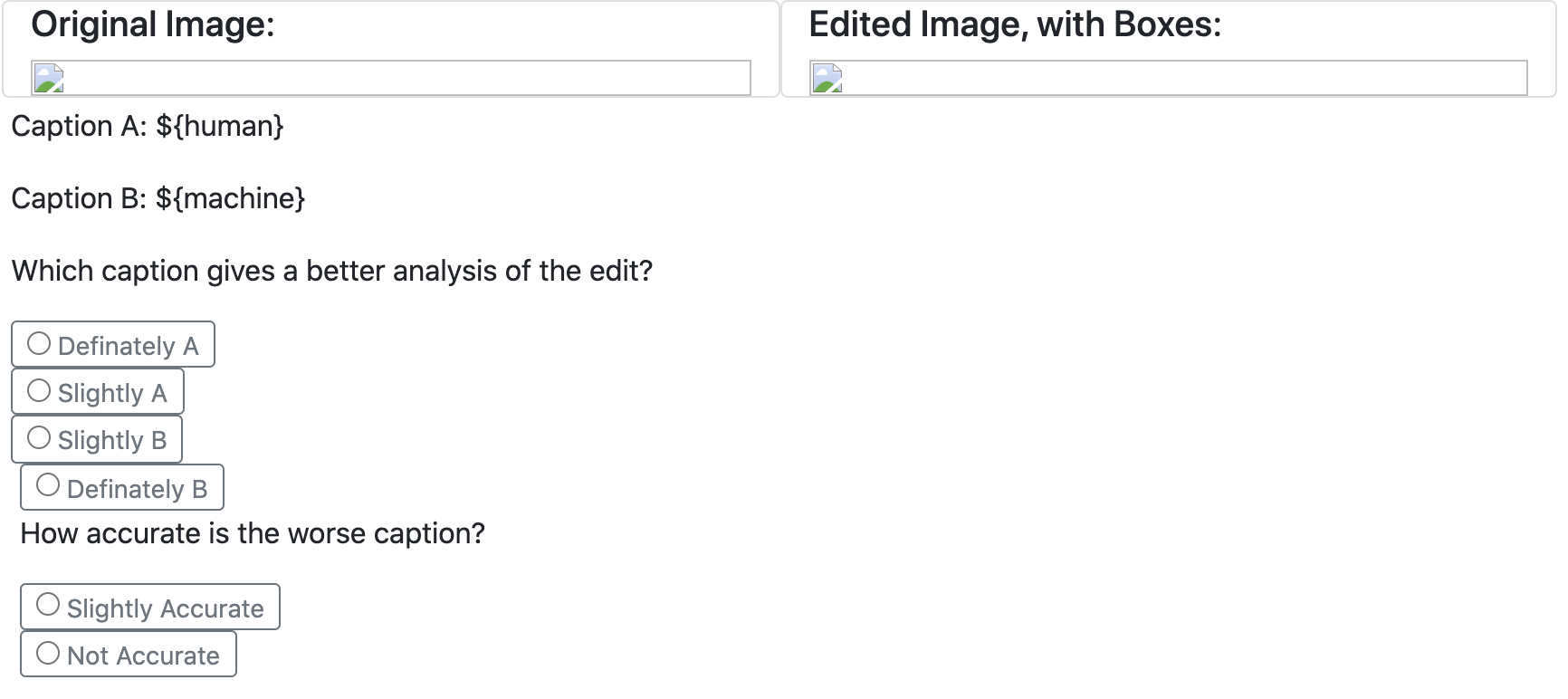}
  \caption{Example of our evaluation process.}
  \label{fig:part7}
\end{figure}

We also show our human evaluation process in Figure \ref{fig:part7}.

\subsection{Additional Annotation Details}

For an image pair (consisting of an image edit and a source image), we 1) ask the annotator to identify and index the main subjects in the image, 2) prime the annotator by asking them to describe the physical change in the image, 3) ask a series of questions for each main person they identified, and 4) ask a series of questions about the image as a whole. For each question we require annotators to provide both an answer to the question and a rationale (e.g. the physical change in the image edit that alludes to their answer). This is critical, as the rationales prevent models from guessing a response such as ``would be harmful'' without providing the proper reasoning for their response. We ask annotators to explicitly separate the rationale from the response by using the word ``because" or ``since" (however, we find that the vast majority of annotators naturally do this, without being explicitly prompted). For the main subjects, we limit the number of subjects to 3. This also mitigates a large variation in workload between image pairs, which was gathered as potentially problematic from annotator feedback. We limit the number of captions per \dimension \space to 3. We find that a worker chooses to provide more than one label for a \dimension \space in only a small proportion of cases, suggesting that usually, one caption is needed to convey all the information about the image edit relating to that \dimension \space.

\begin{figure}[t]
\centering
  \includegraphics[width=0.4\textwidth]{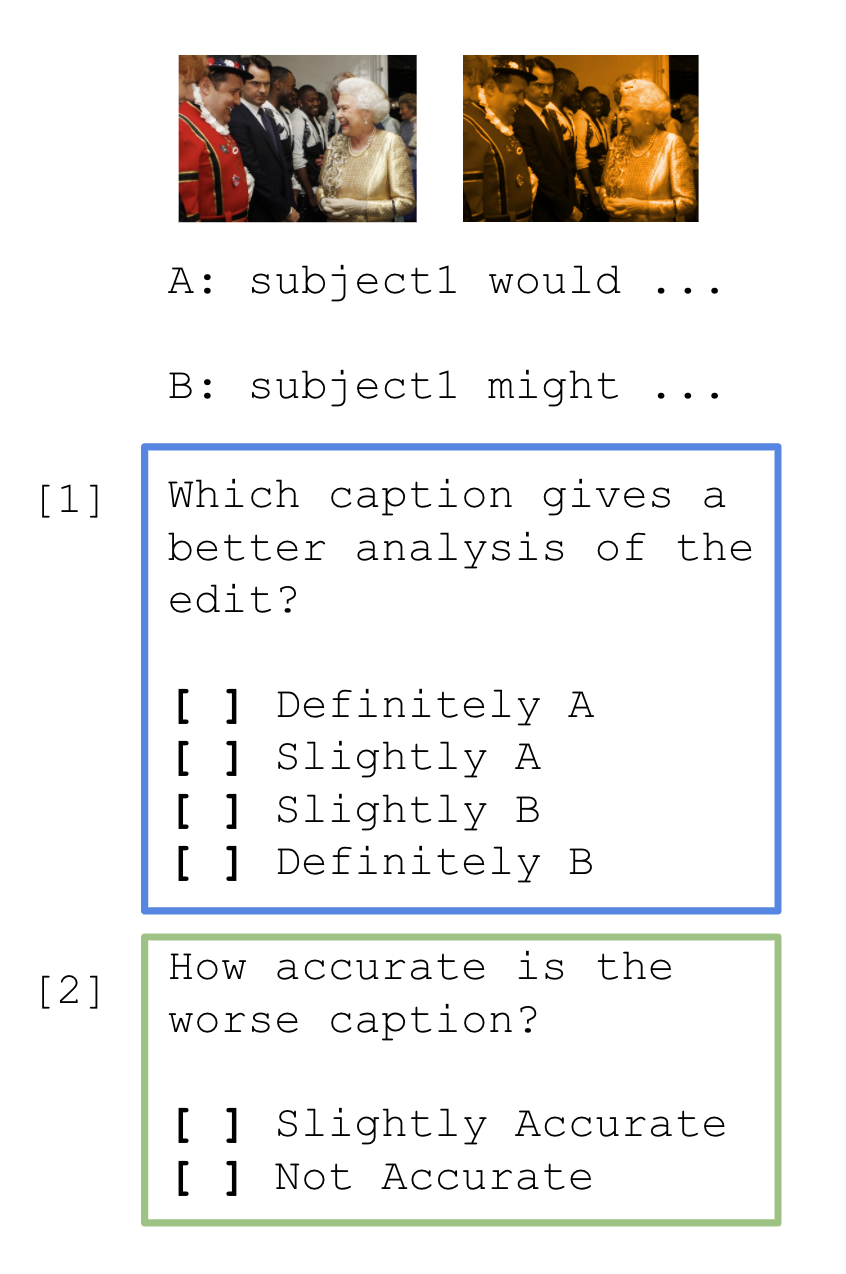}
  \label{fig:humaneval}
  \caption{Our template for human evaluations. Each annotator is shown an edited image, the source image, and is asked to compare a human annotated captions and a machine annotated caption.}
\end{figure}

\subsection{Lexical Analysis}

\begin{table*}[ht]
\centering
\resizebox{\textwidth}{!}{\begin{tabular}{ll||ll|ll|ll|ll|ll}
\multicolumn{2}{c||}{} & \multicolumn{10}{c}{\textbf{Responses}} \\ \cline{3-12}
\multicolumn{2}{c||}{\textbf{Rationales}} & \multicolumn{2}{c|}{intent} & \multicolumn{2}{c|}{implication} & \multicolumn{2}{c|}{disinformation} & \multicolumn{2}{c|}{emotion [of SubjectX]} & \multicolumn{2}{c}{attack [on SubjectX]} \\ \hline
holding & 4.21\% & fun & 4.83\% & public & 3.07\% & movie & 2.93\% & confused & 7.62\% & likes & 3.00\% \\
face & 4.09\% & powerful & 1.13\% & think & 2.12\% & woman & 2.12\% & amused & 4.38\% & hates & 2.21\% \\
wearing & 3.17\% & funny & 1.09\% & man & 1.75\% & new & 1.92\% & embarrassed & 3.88\% & loves & 1.36\% \\
man & 2.64\% & hero & 1.02\% & fun & 1.68\% & game & 1.23\% & upset & 3.50\% & wants & 1.35\% \\
appears & 2.41\% & movie & 1.01\% & disgrace & 1.25\% & real & 1.23\% & proud & 2.61\% & doesn't & 1.31\%
\end{tabular}}
\caption{Lexical statistics. Statistics for each dimension represent omit the rationale, and statistics for the rationale are reported separately.} 
\end{table*}

\textbf{Word-Level Statistics} We analyze the lexical statistics of this dataset. We remove stop words as words such as ``him". We show that different \dimensions \space require different language in their response. In addition, we highlight that many of the rationales involve people, suggesting that understanding social implications is critical to solving this task.

\subsection{Motivation for EMU Task Definition}

We begin by motivating and contextualising our problem. A key insight is that we need to think into the future -- since the task is important but difficult, we aim to structure EMU such that it can help models learn how to understand misinformation (by providing the source image, grounding captions, and additional annotations) without oversimplifying the task.

\textbf{Frames.} We ask models a series of questions about the \textit{what and why} of the image edit. We arrived on these questions by first asking annotators to explain the image edits without prompting. Then, we bucketed the responses into similar categories, motivating us to create questions based on the parts of edits humans naturally focused on. In our task, we consider six open-ended question types -- \textit{physical, intent, implication, emotion [of SubjectX], attack [on SubjectX], and disinformation}. Descriptions of each are in Figure \ref{fig:summary}. Each \dimension \space focuses on a different aspect of the image edit, and is related one-to-one with an open-ended question $\mathbf{q}$. Each question \dimension \space may also reference a specific entity $\mathbf{b}$. In these cases, the answer to the question would differ based on the main subject referred.

\textbf{Labels.} For each $\boldsymbol{q}$, we ask models to provide both a classification label $\mathbf{l}$ and a generated answer (response $\mathbf{y}$ and rationale $\mathbf{r}$) for a given image edit. Visual misinformation is not a closed form problem -- the potential label-space and responses for an malicious edit are ever-changing with recent events. Thus, we suggest that models need to produce a generated answer. However, we also want models to go beyond simple answering -- we want them to answer \emph{for the right reasons}, in an explainable way. Thus, we require models to generate a \emph{rationale} explaining why its answer is true. For example, a good rationale explains that the perception of \subjectone~could be injured because a gun was added to \subjectone's hand. Our evaluation recruits human raters to compare generated answers and rationales $\boldsymbol{y}$/$\boldsymbol{r}$ to those written by annotators. To account for the current difficulty of evaluating generation, we include a binary classification label $\mathbf{l}$ for each of the ``why" answers to allow for a simple checkpoint evaluation metric of model progress. 

\textbf{Grounding.} Each explanation is grounded to bounding boxes $\mathbf{a}_i$ of the people in the edited image. Similar to past work in vision-and-language \cite{Zellers2019FromRT}, annotators write captions that refer to the bounding box (for example, \subjectone would be angry). This allows precise reference in visually complex edits.

\textbf{Additional annotations.} Finally, we provide annotators for bounding boxes of \textit{introduced} and \textit{modified} regions in edited images. These bounding boxes provide the \textit{syntax} of the change in a machine digestible format (bounding boxes + labels). We conduct initial exploration of the empirical benefit of these labels in our modeling section.

\end{document}